\documentclass{article} 
\usepackage{iclr2021_conference,times}

\usepackage{amsmath,amsfonts,bm}

\def\eqref#1{equation~\ref{#1}}

\def\1{\bm{1}}

\DeclareMathAlphabet{\mathsfit}{\encodingdefault}{\sfdefault}{m}{sl}
\SetMathAlphabet{\mathsfit}{bold}{\encodingdefault}{\sfdefault}{bx}{n}

\usepackage{times}
\usepackage{latexsym}
\usepackage{enumitem}
\usepackage{graphicx}
\usepackage{flushend}
\usepackage{marvosym}
\usepackage{hyperref}
\usepackage{wrapfig}

\usepackage{url}

\iclrfinalcopy

\title{Language Acquisition is Embodied,\\ Interactive, Emotive: a Research Proposal}

\author{Casey Kennington \\
Department of Computer Science\\
Boise State University\\
Boise, Idaho, U.S.A. \\
\texttt{caseykennington@boisestate.edu} 
}

\begin{document}

\maketitle

\begin{abstract}
Humans' experience of the world is profoundly multimodal from the beginning, so why do existing state-of-the-art language models only use text as a modality to learn and represent semantic meaning? In this paper we review the literature on the role of embodiment and emotion in the interactive setting of spoken dialogue as necessary prerequisites for language learning for human children, including how words in child vocabularies are largely concrete, then shift to become more abstract as the children get older. We sketch a model of semantics that leverages current transformer-based models and a word-level grounded model, then explain the robot-dialogue system that will make use of our semantic model, the setting for the system to learn language, and existing benchmarks for evaluation. 
\end{abstract}

\section{Introduction}

\cite{Smith2005} showed that babies' experience of the world is profoundly multimodal: babies live in a physical world full of rich regularities that organize perception, action and thought. Babies explore the world in non goal-oriented ways, and babies learn in a social world to learn a shared linguistic communicative system that is symbolic. Indeed, a growing body of literature including child development, psychology, linguistics, and computational linguistics makes a strong case that the process of language learning (indeed, general human cognition) is embodied, interactive, and enacted \citep{Pulvermuller1999,lakoff1999philosophy,Barsalou2008,johnson2008meaning,Smith2009ObjectsIS,di2018linguistic,Bisk2020}. I argue that the \emph{setting} (i.e., where and how the learning takes place) and stages of \emph{progression} of how language is learned matters for holistic knowledge of semantic meaning, which has implications for how language is modeled computationally, especially in light of the fact that most language models, including recent transformer-based models like BERT \citep{Devlin2018} and GPT-3 are derived abstractly from adult-written text. 

\begin{wrapfigure}{r}{0.55\textwidth} 
\centering
  \includegraphics[width=0.55\textwidth]{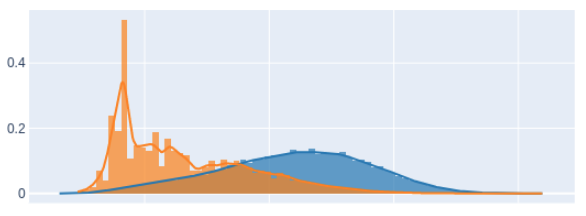}
  \caption{Average Age-of-Acquisition ratings for each entry in a subset of the WordNet dictionary: average ratings for word entries (blue) are higher than ratings for words in the definitions for those entries (orange).  \label{fig:aoadict}}
\end{wrapfigure} 

Basic evidence that progression of learning matters is found in age-of-acquisition (AoA) datasets where known words are annotated with the average age when children are able to produce those words. \cite{Kuperman} presented ratings for over 30,000 English words (including nouns, verbs, and adjectives). For example the word \emph{red}'s rating is 3.68 (i.e., 3 years + 0.68 towards the 4th year), and \emph{abandon} is 8.32. Taking definitions from WordNet \citep{Miller1995} for words that exist on the AoA dataset (totaling 26,919 words), the average AoA age for the words is 11 years (std 3.04), whereas the average AoA age for all of the words in the definitions is 6.59 (std 2.67). This is further illustrated in Figure~\ref{fig:aoadict}, showing that words that make up definitions in the WordNet dictionary are learned much younger. This seems trivial; obviously words that are learned earlier in life are used to learn the meaning (or at least the definition) of words later in life, but language models and the data they leverage do not take this progression into account. 


\cite{Vincent-Lamarre2016} gave this deeper consideration and found that recursively removing all words that are reachable by their definitions, but that do not define any further words, a dictionary can be reduced to 10\% of its size. This can further be reduced to a strongly connected subset of words that all other words can be defined from, which comprises only about 1\% of the dictionary.  This is important because while many words are defined by other words, there is a core subset of words that cannot be defined by other words (e.g., what is the dictionary definition of the word \emph{red}?), but rather must be experienced directly, otherwise meaning of all words are completely ungrounded; ``just strings of meaningless symbols (defining words) pointing to meaningless symbols (defined words)" which is precisely what the symbol grounding problem is \citep{harnard:grounding}.\footnote{Note that the author of \cite{harnard:grounding} is an author of \cite{Vincent-Lamarre2016}, which makes these claims.} \cite{Vincent-Lamarre2016} showed that the ``core" words, from which all other words are eventually defined, are learned earlier in life and they are more \emph{concrete}; i.e., they are words that denote physical, tangible objects and proprioperceptive embodied states. This isn't to say that humans don't have the capacity for abstraction without concrete experience. Indeed, all words are to some degree abstract because they make an abstract categorization, even ones that directly denote perceptual experience \citep{Harnad2017}. But given the literature cited above, no human would likely arrive at abstract thought without something to abstract over, i.e., words that denote concrete, physical entities. 

\paragraph{Requirements} Despite important advances for natural language processing (NLP) tasks and applications, it is clear that models trained only on text are missing critical semantic information (see, for example, \cite{Rogers2020} which reviews relevant literature). These kinds of text-only models make an \emph{abstractness assumption} because the only ``context" they make use of is lexical context (i.e., words are only ``defined" by other words), whereas a holistic model of meaning requires lexical, embodied (including emotion, see \cite{lane2002cognitive,Moro2020}), perceptual (i.e., connected to the world--symbol grounding), and interactive (i.e., conversational grounding \citep{Clark1996}) context in the language learning process. 

\section{Research Proposal}

The goal of this research project is to work towards a model of semantic meaning that handles concrete and abstract meaning acquisition, encodes emotion, and is learned in a setting similar to that of a child: embodied, spoken interaction. 

\paragraph{Embodiment} The semantic model needs to be housed in a physical body that can perceive and act in the world. We opt to use robotic platforms, beginning with Anki Cozmo which has been shown to be suitable for the setting of first-language acquisition because people perceive Cozmo to have a young age \citep{Plane2018}, though clearly Cozmo is nothing like a real child in its sensory capabilities or afforances. Cozmo's perceptual abilities include camera input, we add an external microphone, and tracking of internal state variables (e.g., lift height, head angle, wheel speed) and Cozmo's abilities for action include driving forward and backward, turning, lifting and lowering a small lift arm that can move specific types of objects, as well as up and down movement of the head and speech synthesis with a young-sounding voice.

\paragraph{Interaction} The setting for the robot to learn language is face-to-face spoken, interactive dialogue which is the basic setting where humans learn their first language \citep{Fillmore1975}. This \emph{situated} setting requires a spoken dialogue system (SDS) that is well-suited for robots.  Following \cite{Kennington2020}, such a robot-ready SDS must be (1) \emph{modular} so it can integrate with various robot modules, (2) \emph{multimodal} the semantic model should incorporate perception (including proprioperception), (3) \emph{distributive} so robot modules and SDS modules can communicate with each other across distributed hardware, (4) \emph{incremental} so processing can happen quickly and immediately, and (5) \emph{aligned} in that sensors and actions must be synchronized temporally. The incremental requirement is crucial: the semantic model must not wait for full, grammatical utterances; rather, it should process by word (or sub-word) increments because humans process spoken input in real-time \citep{Eberhard1995}, though many models of semantics require full, sentence-level input. 

\paragraph{Emotion} The meanings of many words have emotion as part of their connotation \citep{lane2002cognitive} and that emotion plays a role in the meaning of abstract words \citep{Vigliocco2014}. Following \cite{Moro2020}, similar to semantic word meaning, emotions can be viewed as on a continuum between abstract and concrete; abstract according their lexical categories (e.g., \emph{happiness}, \emph{fear}, \emph{anger}) distributed with text \citep{barrett2017emotions}, and concretely through \emph{affect} which is a biological system and a fundamental part of embodiment \citep{Vigliocco2014}. In contrast to abstract emotion concepts, affect is a more basic underpinning for emotion, ranging from unpleasant to pleasant (valence) and from agitated to calm (arousal), which, like vision, is something that could potentailly be grounded into if a model exists. We use the model we introduced in \citep{Kennington-RSS-19} as a proxy for affect as it maps from robot behaviors to a distribution over 16 affects that were labeled by humans who observed the behaviors for affective display. 

\paragraph{Open Questions} This research highlights some important questions relating to semantic meaning of language, how it is learned, the role of emotion in meaning (and language acquisition) which has implications for NLP, robotics, human-robot interaction, and artificial intelligence applications. Specifically, we ask the following questions:

\begin{itemize}
    \item What semantic model fulfills the requirements of being grounded, can learn meanings of words with only a few examples (i.e., fast-mapping as children can do) or directly from explanation, and can be learned through interaction with others and with the physical environment?
    \item How can concrete and abstract meaning be learned and represented in that model?
    \item Can affect and emotion help reconcile the concrete and abstract meaning learning and representations?
\end{itemize}

\paragraph{Model Sketch} Two models that inspire our proposed model (though there are many other vision-lanuage models) are (1) VilBERT \citep{Lu2019}, a dual-transfomer architecture that brings together textual embeddings and images for image description generation, more recently leveraged for visual dialogue \citep{visdial_bert}, and (2) the \emph{words-as-classifiers} (WAC) model \citep{Kennington2015_acl}, a simple grounded modal that amounts to a binary classifier for each word, each classifier yields a ``fitness" score, given a representation of a visual object and word in question. \cite{Schlangenetal2016} use the WAC model with images of real objects in a reference resolution task, using vectorized representations of objects from a convolutional neural network trained on imagenet data. WAC has been used as a grounded model for modalities beyond vision; for example, \cite{Moro2020} grounded WAC into low-level affect predictions which included robot internal states and audio representations.

Both models have their advantages. VilBERT is transformer-based and leverages BERT for representing language, making it robust to various language input. Being a word-level model, WAC has the advantage of being useable in an incremental SDS setting--a requirement for robot-ready SDS--and can learn fitness scores with only a few training examples. For example, \cite{McNeill2020} recently used WAC in an interactive language acquisition study using WAC as the semantic model on the Cozmo robot with human participants; WAC was able to quickly learn a handful of vocabulary words despite the short interactions. Both models have disadvantages. VilBERT, like most neural models, is data hungry and uses BERT which is trained on text, yet children learn interactively and often with only a small set of examples. Moreover, VilBERT is currently strictly designed to model the visual modality taken from images. WAC suffers from two strong assumptions. First, that all words are trained and applied independently from each other (making WAC's composition strategy quite limiting--it simply multiplies the fitness scores together to form sentence-evel scores for referring to objects) and second, that all words fully denote concrete things, despite many words being abstract and therefore do not have physical manifestations (e.g., \emph{utopia} or \emph{beneficial}).

Our current work explores using WAC classifiers as embeddings for the VilBERT model by training WAC classifiers (i.e., simple multi-layer perceptrons) on images using positive and negative image examples for each word, then extracting the coefficients of those classifiers (i.e., concatenate the coefficients for each layer, forming a vector) that we then use as input embeddings for the language side of the VilBERT model. Current results show promise on the visual dialogue benchmark \citep{visdial_bert} that this is a useful addition to the VilBERT model, thereby unifying the two models and overcoming some of the assumptions and shortcomings of each model used alone, but more work is needed for a model that fulfills all of the requirements and works for a robotic platform in a spoken dialogue setting. 

\begin{figure*}
\centering
  \includegraphics[width=1.0\textwidth]{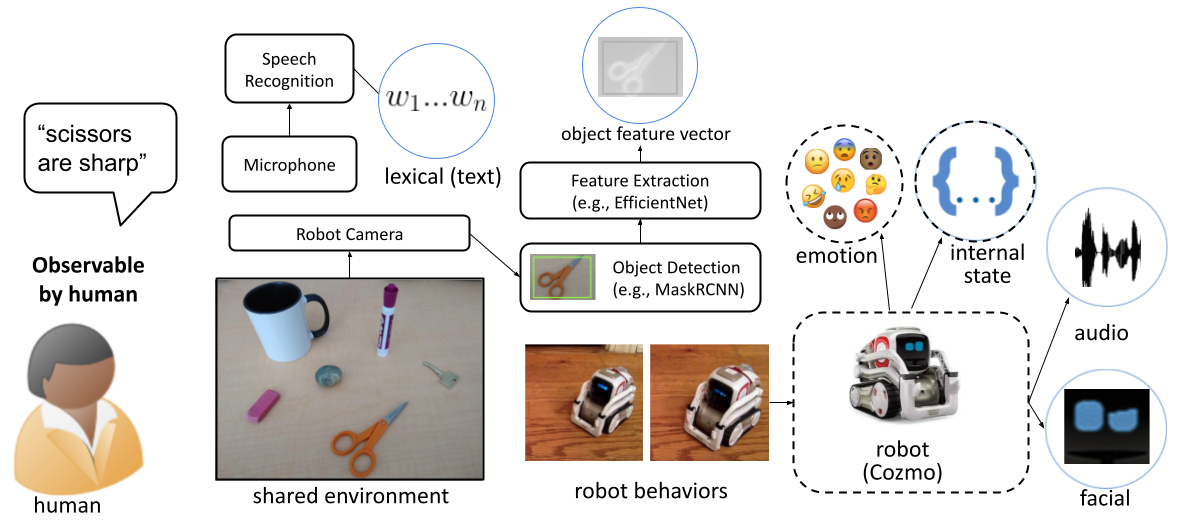}
  \caption{Overview of proposed system. The model takes in 6 modalities (depicted in the circles); modalities come from the user’s speech, shared environment, or robot. Emotion is grounded into by the model, and (including emotional) robot behaviors facilitate the interaction. The user utterance contains concrete and abstract terms.  \label{fig:overview}}
\end{figure*}

\paragraph{System} We will use the system for incremental dialogue described in \cite{Kennington2020}, which has modules for speech recognition, object detection using YOLOv4 \citep{bochkovskiy2020yolov4}, and object feature representation by using the topdropout layer from EfficientNet \citep{Tan2019} to feed into our semantic model. The rrSDS platform has bindings for the Cozmo robot, as well as bindings for OpenDial \citep{Lison2016}, which we will use for dialogue act and robot action decisions. We will use the model described in \cite{Kennington-RSS-19} to map from robot modalities to a distribution over a representation of affect that our model of semantics will ground into. Our system is portrayed visually in Figure~\ref{fig:overview}, including the Cozmo robot.

\paragraph{Evaluation Plan}

We will recruit human participants and task them with interacting with Cozmo first by referring to concrete objects (e.g., \emph{scissors near you}, as done in \cite{McNeill2020}, then task them with engaging in more abstract comparisons or truth claims (e.g., \emph{scissors are sharp}). To gain exposure to larger vocabulary, we will put Cozmo in varied contexts with participants where participants also interact with Cozmo at intervals across a long time span. We hypothesize that this will result in a representation of semantic meaning encoded in a model like VilBERT that has higher fidelity to the setting and circumstances in which human children learn language. We will continue our ongoing work in using this model on known benchmarks (e.g., GLUE \citep{Wang2018}, visual dialogue \citep{visdial_bert}) to compare with existing models that are only trained on text data.

\section{Conclusions}

The process whereby human children learn language is vastly different from the process whereby existing state-of-the-art language models learn language. While current advancements in multimodal language grounding are moving in the right direction, the setting (i.e., situated dialogue) and lack of embodiment still pose a challenge. Addressing these requirements in a single system is by no means low-hanging research fruit; it requires interdisciplinary background in NLP, SDS, and human-robot interaction research, but we believe that the efforts will be beneficial in working towards natural communication with automated systems, including robots.


\bibliography{iclr2021_conference}
\bibliographystyle{iclr2021_conference}


\end{document}